# Integration of Contextual Descriptors in Ontology Alignment for Enrichment of Semantic Correspondence

Eduard Manziuk
*Department of Computer Science*
*of Khmelnytskyi National University*
Khmelnytskyi, Ukraine
eduard.em.km@gmail.com

Oleksander Barmak
*Department of Computer Science*
*of Khmelnytskyi National University*
Khmelnytskyi, Ukraine
barmako@khmnu.edu.ua

Pavlo Radiuk
*Department of Computer Science*
*of Khmelnytskyi National University*
Khmelnytskyi, Ukraine
radiukp@khmnu.edu.ua

Vladislav Kuznetsov
*Department of Theoretical Cybernetics,*
*Taras Shevchenko National University*
*of Kyiv*
Kyiv, Ukraine
kuznetsow.wlad@gmail.com

Iurii Krak
*Laboratory of Communicative*
*Information Technologies, V.M.*
*Glushkov Institute of Cybernetics*
Kyiv, Ukraine
iurii.krak@knu.ua

Sergiy Yakovlev
*Mathematical Modelling and Artificial*
*Intelligence Department, National*
*Aerospace University "Kharkiv*
*Aviation Institute"*
Kharkiv, Ukraine
s.yakovlev@khai.edu

*Abstract* — **This paper proposes a novel approach to semantic ontology alignment using contextual descriptors. A formalization was developed that enables the integration of essential and contextual descriptors to create a comprehensive knowledge model. The hierarchical structure of the semantic approach and the mathematical apparatus for analyzing potential conflicts between concepts, particularly in the example of "Transparency" and "Privacy" in the context of artificial intelligence, are demonstrated. Experimental studies showed a significant improvement in ontology alignment metrics after the implementation of contextual descriptors, especially in the areas of privacy, responsibility, and freedom & autonomy. The application of contextual descriptors achieved an average overall improvement of approximately 4.36%. The results indicate the effectiveness of the proposed approach for more accurately reflecting the complexity of knowledge and its contextual dependence.**

I. INTRODUCTION

Ontology alignment is a critical task in the integration and exchange of knowledge across different systems and domains. The goal of alignment is to identify semantic correspondences between entities in different ontologies, to ensure consistent data representation and exchange. Numerous studies have been dedicated to developing effective ontology alignment methods, particularly those using machine learning techniques and knowledge embeddings.

However, existing approaches have certain limitations and drawbacks, such as difficulty in generalizing to new domains, reliance on additional data or expert annotations, and challenges in quality assessment and consideration of the semantic and structural information of ontologies [1–3]. A promising direction is the development of universal methods for heterogeneous ontologies, improving quality assessment, and managing ontology evolution and alignments.

This study proposes the application of contextual descriptors in the semantic ontology alignment method. By distinguishing between essential and contextual descriptors and integrating them, a comprehensive knowledge model can be created that considers both the internal structure of knowledge bases and external contextual factors that influence the interpretation and application of knowledge. This approach allows for a more accurate representation of the complexity and multidimensionality of knowledge, particularly when working with heterogeneous data and integrating information from various sources.

The main contributions of the study are as follows:

• a new approach to ontology alignment is proposed by integrating contextual descriptors in the semantic determination of the correspondence of substantive categories;

• a formalization of the application of contextual descriptors in the semantic approach is developed;

• a hierarchical structure of the semantic approach components with a division into essential and contextual descriptors is proposed;

• the advantages of integrating essential and contextual descriptors for creating a comprehensive knowledge model capable of adapting to real conditions and ensuring relevant information use in various contexts are presented.

The structure of the paper is as follows. The introduction defines the problem of ontology alignment, the limitations of existing methods, and the motivation for using contextual descriptors. The literature review covers modern approaches to ontology alignment, their limitations, and prospects. Section III details the proposed approach, specifying the application of essential and contextual descriptors and formalizing the method. The methodology describes the hierarchy of descriptors, their integration, and the ontology alignment algorithm. Section IV presents the experimental evaluation, including data and results compared with other methods. The discussion analyzes the results, advantages, limitations, and directions for future research. The conclusions summarize the contributions of the study and its significance for future work.

## II. RELATED WORKS

One of the leading approaches is the application of representation learning techniques. Such research is presented in [4], where a method with the integration of these techniques, retrofitting, and autoencoders was proposed. In [5, 6], VeeAlign systems and the enhancement of traditional methods through machine learning, , and semantic networks, respectively, were introduced. In [7], transfer learning with BERT was used for ontology alignment. In combination with symbolic methods [8], the symbolic FCA-Map method was integrated with the Siamese BERT vector model for biomedical ontology alignment. The formalization of semantic meanings of substantive categories plays a significant role in practical applications [9–11].

The consideration of ontology semantics carried out in [12], which combines embedded knowledge from knowledge graphs and ontologies, considering the semantics of class hierarchies. In [13–15], ontology alignment in specific domains, such as pharmaceuticals, the Internet of Things, and industrial maintenance, is discussed.

Studies [3, 16, 17] focus on methodological aspects, including benchmark alignment learning, reviews of existing methods, and analysis of current systems. Papers [18–20] provide benchmarks, methodologies for maintaining alignment validity, and reviews of cross-linguistic and graph-based approaches.

Studies [1, 21–23] propose various structural methods, such as graph embeddings, negative sampling, concept category merging, and improved accuracy metrics. [24, 25] present new structural approaches based on embedded representations and indirect cross-linguistic alignment.

The key limitations and challenges identified in the analyzed works are the difficulties of generalization and application to new domains and heterogeneous datasets. There is a dependence on additional knowledge, expert annotations, or partially aligned data. There is a need to improve quality assessment, considering the structural and semantic characteristics of ontologies. Managing ontology evolution and maintaining valid alignments remain complex tasks, with performance constraints on certain data types and when aligning heterogeneous ontologies.

Promising directions include developing universal methods for diverse and heterogeneous ontologies, improving quality assessment, considering structural and semantic information, and effectively managing ontology evolution and alignments. Combining symbolic, semantic, and domain-specific approaches, which are oriented towards specific domains, remains promising but requires additional effort to address the identified limitations.

Thus, *the main goal of the research* is to improve the accuracy of semantic correspondence to ensure the relevant use of knowledge in different contexts. This goal is proposed to be achieved by developing an approach to integrating contextual descriptors into the ontology alignment process.

Therefore, *the goal of the research* is to develop an approach to integrating contextual descriptors into the ontology alignment process to improve the accuracy of semantic correspondence and ensure the relevant use of knowledge in various contexts.

## III. APPROACH TO THE APPLICATION OF CONTEXTUAL DESCRIPTORS IN THE SEMANTIC DETERMINATION OF CORRESPONDENCE OF ONTOLOGY SUBSTANTIVE CATEGORIES

The method of semantic determination of content category correspondence for comparing knowledge bases in the form of ontologies or schemas can be improved for the specific case of studying the correspondence between the trustworthiness AI ontology and a structured domain based on a data corpus. In this context, accounting for data diversity is key, requiring the integration of knowledge from different sources to achieve the most complete understanding.

At the initial stage, it is necessary to distinguish between essential and contextual descriptors to improve the analysis and comparison process. Essential descriptors reflect the internal descriptive structures and relationships between elements in the knowledge base. They are responsible for how knowledge is organized and represented at the formal level. These descriptors include aspects such as hierarchical structures, types of connections between descriptive structures, as well as the basic properties that define entities.

Contextual descriptors, on the other hand, focus on situational or external aspects that affect the use or interpretation of knowledge. They cover factors such as the purpose of information use, specific conditions under which the knowledge was obtained, or cultural and social contexts that may influence the perception and interpretation of knowledge. For example, the same entity may have different meanings or priorities depending on the context in which it is used. Contextual descriptors play a key role in how knowledge is applied in different situations and can influence the outcome of analysis or comparison.

The distinction between essential and contextual descriptors allows for a more accurate representation of the complexity and multidimensionality of knowledge, which is particularly important when working with heterogeneous data and integrating knowledge from various sources. This distinction contributes to more precise and nuanced comparisons, enabling the identification of subtle differences and similarities between entities that may seem similar at first glance but may have different meanings or importance in different contexts.

Within the semantic approach, the crucial step is not only the distinction but also the integration of essential and contextual descriptors to create a comprehensive knowledge model. Essential descriptors provide a foundation for organizing knowledge by creating clear hierarchies and connections between entities. However, without considering contextual descriptors, such models may be limited in their ability to adapt and apply in real situations.

Contextual descriptors add important layers of meaning that consider real-world conditions and variables that may influence the interpretation of knowledge. For example, the same entity in different cultural or social contexts may have different meanings or be perceived in varying ways. Including contextual descriptors allows the model to adapt to these variables and provides the opportunity for deeper understanding.

Within the semantic approach, contextual descriptors also help identify potential sources of semantic heterogeneity that may arise when integrating knowledge from different sources. This is particularly relevant when working with heterogeneous data, where it is necessary to integrate information from different systems, each of which has its unique contextual characteristics. Knowledge of these contextual factors allows for more effective alignment and merging of information, reducing the risk of losing meaning or distorting meanings.

The overall structural hierarchy is presented in Fig. 1

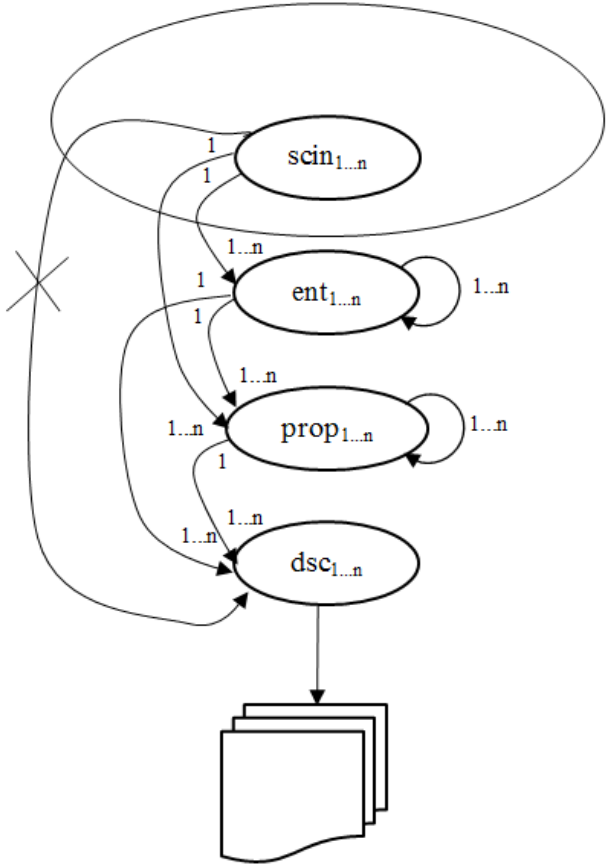

Fig. 1. General structural hierarchy for the semantic determination of correspondence of substantive categories.

In the semantic approach to knowledge analysis, contextual descriptors serve as an integral element that enriches the model and allows for greater accuracy and correspondence to the real world. This contributes to the creation of more flexible and adaptive knowledge systems capable of effectively operating in diverse conditions and environments. Through this approach, not only can entities be better understood, but their correct and relevant use in different contexts can also be ensured. Subsequently, entities can be generalized into such structures as information scope. The primary purpose of building a hierarchical architecture is to formally and structurally generalize knowledge of the subject domain.

To represent essential and contextual descriptors in a formalized way, relationships can be used that represent entities, their properties, and descriptors, which are divided into essential and contextual.

Let's formalize the structural hierarchy of the semantic approach, starting with entities. The overall structure of information scope within this study will remain unchanged, as the formation of information scope is carried out without the direct use of descriptors.

Let's define a relationship for entities. Let $Ent = \{ent_1, ent_2, \ldots, ent_n\}$ – the set of entities, and $Prop = \{prop_1, prop_2, \ldots, prop_n\}$ – the set of properties. Each entity can be represented by the relationship $Rel_{prop} = (Ent, Prop, Val_{prop})$, where $Val_{prop} = \{val_{prop_1}, val_{prop_2}, \ldots, val_{prop_n}\}$ – the set of property values for each entity.

The definition of the relationship is carried out within a strict hierarchy. Although in general, to ensure the reproduction of language heterogeneity, an entity can be defined using a non-strict hierarchy, that is, it can be formed using omissions of the strict hierarchy. However, for the sake of simplifying the presentation, a strict hierarchy will be applied.

The relationship $Rel_{prop}$ describes the connection between an entity, a property, and its value:

$$Rel_{prop} = \begin{cases} (ent_i, prop_j, val_{prop_{ij}}) | \\ \forall ent_i \in Ent, \forall prop_j \in Prop, \\ \forall val_{prop_{ij}} \in Val_{prop} \end{cases} \quad (1)$$

Relationship for properties based on essential descriptors. Let $Dsc_{essent} = \{dsc_{essent_1}, dsc_{essent_2}, \ldots, dsc_{essent_n}\}$ – the set of essential descriptors. The relationship of properties using essential descriptors can be represented as $Rel_{dsc\_essent} = (Prop, Dsc_{essent}, Val_{dsc\_essent})$, where $Val_{dsc\_essent} = \{val_{dsc\_essent_1}, val_{dsc\_essent_2}, \ldots, val_{dsc\_essent_n}\}$ – the set of essential descriptors values for each property.

The relationship $Rel_{dsc\_essent}$ describes the connection between a property, a essential descriptor, and its value:

$$Rel_{dsc\_essent} = \begin{cases} (prop_i, dsc_{essent_j}, val_{dsc_{essent_{ij}}}) | \\ \forall prop_i \in Prop, \forall dsc_{essent_j} \in Dsc_{essent}, \\ \forall val_{dsc_{essent_{ij}}} \in Val_{dsc_{essent}} \end{cases} \quad (2)$$

Relationship for properties based on contextual descriptors.

Let $Dsc_{contex} = \{dsc_{contex_1}, dsc_{contex_2}, \ldots, dsc_{contex_n}\}$ – the set of contextual descriptors. The relationship of properties using contextual descriptors can be represented as $Rel_{dsc\_context} = (Prop, Dsc_{contex}, Val_{dsc\_contex})$, where $Val_{dsc\_contex} = \{val_{dsc\_contex_1}, val_{dsc\_contex_2}, \ldots, val_{dsc\_contex_n}\}$ – the set of contextual descriptors values for each property.

The relationship $Rel_{dsc\_context}$ describes the connection between a property, a contextual descriptor, and its value:

$$Rel_{dsc\_contex} = \begin{cases} \left(prop_i, dsc_{contex_j}, val_{dsc_{contex_{ij}}}\right) | \\ \forall prop_i \in Prop, \forall dsc_{contex_j} \in Dsc_{contex}, \\ \forall val_{dsc_{contex_{ij}}} \in Val_{dsc_{contex}} \end{cases} \quad (3)$$

In fact, a descriptor is divided into two types: a factual descriptor and a contextual descriptor. That is, further specification of certain generalizing descriptive constructions is carried out (Fig. 2).

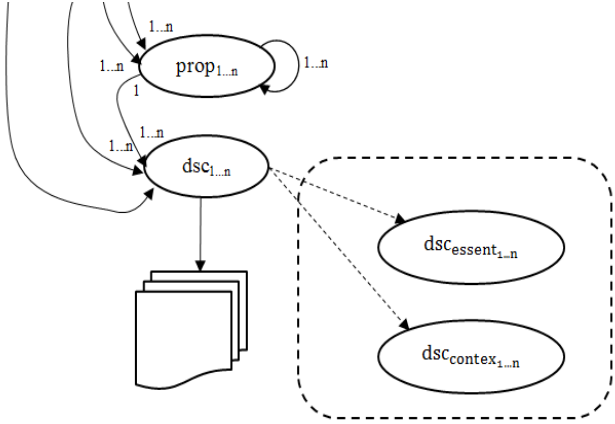

Fig. 2. Division of descriptors into essential and contextual components in the semantic determination of correspondence.

To obtain the set of property values and descriptors, appropriate projections are applied.

Thus, to obtain all property values for a specific entity $ent_i$, the projection operation on the corresponding set is applied:

$$\pi_{Prop, Val_{prop}}\left(Rel_{prop}\right) = \left\{\left(prop_j, val_{prop_{ij}}\right) | ent_i \in Ent\right\} \quad (4)$$

Similarly, to obtain all essential descriptor values for a property $prop_i$:

$$\pi_{Dsc_{essent}, Val_{dsc\_essent}}\left(Rel_{dsc\_essent}\right) = \\ = \left\{\left(dsc_{essent\,j}, val_{dsc_{essent_{ij}}}\right) | prop_i \in Prop\right\} \quad (5)$$

Also, to obtain all contextual descriptor values $dsc_{essent\,j}$ for the property:

$$\pi_{Dsc_{contex}, Val_{dsc\_contex}}\left(Rel_{dsc\_contex}\right) = \\ = \left\{\left(dsc_{contex\,j}, val_{dsc_{contex_{ij}}}\right) | prop_i \in Prop\right\} \quad (6)$$

The next step is to combine essential and contextual descriptors

To integrate essential and contextual descriptors, the union of $Rel_{dsc\_essent}$ and $Rel_{dsc\_contex}$:

$$Rel_{dsc} = Rel_{dsc\_essent} \bowtie Rel_{dsc\_contex} \quad (7)$$

Thus, the union of relationships by property creates a new relationship that includes both essential and contextual descriptors.

To compare entities by properties, appropriate matches are selected:

$$\sigma_{val_{prop_{ij}} = val_{prop_{km}}}\left(Rel_{prop}\right) = \\ = \left\{(ent_i, ent_k) | val_{prop_{ij}} = val_{prop_{km}}\right\} \quad (8)$$

This allows identifying entities that have the same values for a particular property.

Similarly, for contextual descriptors:

$$\sigma_{val_{dsc\_contex_{ij}} = val_{dsc\_contex_{km}}}\left(Rel_{dsc\_contex}\right) = \\ = \left\{(prop_i, prop_k) | val_{dsc_{contex_{ij}}} = val_{dsc_{contex_{km}}}\right\} \quad (9)$$

and essential descriptors:

$$\sigma_{val_{dsc\_essent_{ij}} = val_{dsc\_essent_{km}}}\left(Rel_{dsc\_essent}\right) = \\ = \left\{(prop_i, prop_k) | val_{dsc_{essent_{ij}}} = val_{dsc_{essent_{km}}}\right\} \quad (10)$$

For the generalized description of a property based on essential and contextual descriptors, a new relationship that combines all descriptors can be formed:

$$Rel_{dsc}(prop_i) = \pi_{Prop, Dsc_{essent}, Dsc_{contex}}\left(Rel_{dsc_{essent}}\right) \\ \pi_{Prop, Dsc_{essent}, Dsc_{contex}}\left(Rel_{dsc_{contex}}\right) \quad (11)$$

This relationship contains all the information about the property $prop_i$, which is used for further comparison or analysis.

To analyze the differences between entities, the difference between relationships can be applied:

$$Rel^{diff}_{prop_{ent_i - ent_k}} = Rel_{prop}(ent_i) - Rel_{prop}(ent_k) \quad (12)$$

This relationship represents the difference between the properties of entities $ent_i$ and $ent_k$.

Of course, the same approach is used to analyze essential and contextual descriptors.

These expressions allow for structured and systematic processing of data represented as entities with properties and essential and contextual descriptors, enabling their analysis and comparison within descriptive domains.

## IV. DESCRIPTION OF THE EXPERIMENT

To determine the necessity of expanding descriptor types, we will conduct an experimental study of ontology alignment and analyze the effectiveness of applying contextual descriptors based on works [26, 27]. Let's describe the process of conducting experimental research.

1. Separation of descriptors into essential and contextual.

As an example, we present a table of descriptor distribution for the concept Responsibility, which is one of the key ethical principles for the implementation of artificial intelligence and is obtained based on the analysis of domain [26], which contains 84 sources.

TABLE I. PROPERTIES, DESCRIPTORS, AND TYPES OF THE CONCEPT RESPONSIBILITY

| Property | Descriptor | Type | Number of Sources |
|---|---|---|---|
| AI Decision Accountability | Legal frameworks for AI decisions | Formal | 75 |
| | Internal control mechanisms | Formal | 72 |
| | Public discussion of accountability | Contextual | 65 |
| | Stakeholder involvement in AI accountability | Contextual | 60 |
| Legal Liability | Contracts and legal obligations | Formal | 80 |
| | Legislation regulating AI | Formal | 70 |
| | Cultural differences in legal liability | Contextual | 65 |
| | Distribution of liability among actors | Contextual | 60 |
| Ethical Oversight | Internal ethics committees | Formal | 72 |
| | Audits of ethical compliance | Formal | 68 |
| | Influence of civil organizations | Contextual | 60 |
| | Public opinion influence on ethics committees | Contextual | 62 |
| Transparency in Decision-making | Publicly available AI decision documentation | Formal | 68 |
| | Independent audits of AI decision processes | Formal | 66 |
| | Public expectations for transparency | Contextual | 63 |
| | Cultural expectations of AI transparency | Contextual | 64 |
| Risk Assessment | AI risk management audits | Formal | 73 |
| | Risk forecasting for societal impact | Contextual | 64 |
| | Impact of AI risks on public trust | Contextual | 65 |
| Integrity | Code of Conduct for AI Developers | Formal | 70 |
| | Internal Audit Processes | Formal | 69 |
| | Ethical Compliance Documentation | Formal | 68 |
| | Cultural Norms Influencing Integrity | Contextual | 65 |
| | Stakeholder Trust in Integrity Practices | Contextual | 63 |
| Awareness | Mandatory Ethics Training Programs | Formal | 72 |
| | Ethics Compliance Checks | Formal | 70 |
| | Public Awareness Campaigns | Contextual | 60 |
| | Influence of Social Media on Awareness | Contextual | 58 |
| Diversity and Inclusion | Diversity Hiring Policies | Formal | 65 |
| | Community Engagement Initiatives | Contextual | 60 |
| | Cultural Sensitivity in AI Applications | Contextual | 59 |
| Educational Responsibility | Curriculum Standards for AI Ethics | Formal | 67 |
| | Public Perception of Educational Initiatives | Contextual | 66 |
| Attribution of Responsibility | Legal Frameworks Defining Accountability | Formal | 74 |
| | Contracts Specifying Roles and Liabilities | Formal | 68 |
| | Public Expectations of Accountability in AI | Contextual | 62 |

Criteria that formed the basis for identifying descriptors and dividing them by types:
Descriptor classification criteria:
Essential descriptors:
- Objectively measurable or clearly defined characteristics.
- Easily documented and standardized.
- Usually have a legal or regulatory basis.
- Can be unambiguously verified or confirmed.
- Often associated with specific processes, procedures, or structures.

Contextual descriptors:
- Depend on social, cultural, or situational context.
- May vary depending on interpretation or perception.
- Often reflect implicit or intangible aspects.
- May change over time or depending on social trends.
- Usually associated with public opinion, expectations, or perception.

2. To calculate similarity, we generalize the calculation representation in the following form:

$$S = \frac{\sum_{i=1}^{n_f} s_i^f log\left(1+src_i^f\right) + \sum_{j=1}^{n_c} s_i^c log\left(1+src_j^c\right)}{\sum_{i=1}^{n_f} log\left(1+src_i^f\right) + \sum_{j=1}^{n_c} log\left(1+src_j^c\right)}, \quad (13)$$

where $n_f, n_c$ – number of essential and contextual descriptors respectively;

$s_i^f, s_i^c$ – similarity of the i-th essential and j-th contextual descriptor;

$src_i^f, src_j^c$ – number of sources for the respective descriptors.

Given the small number of sources, an expert approach was used to establish correspondence based on individual assessment with subsequent averaging. It allows for consideration of deep understanding of the subject area and nuances that may be missed by automatic methods.

Performing initial alignment using only essential descriptors with the application of lexical analysis methods. Then, calculating the base percentage of matches.

3. The next step is to supplement each concept with corresponding contextual descriptors and conduct a relevant analysis regarding the similarity obtained using the combined method (essential + contextual descriptors) $S_{f,c}$ and the similarity obtained by the basic method (only essential descriptors) $S_f$.

$$Imp = \frac{S_{f,c} - S_f}{S_f} \quad (13)$$

V. EXPERIMENTAL RESEARCH

To determine the need for expanding the types of descriptors, an experimental study was conducted on the alignment of ontologies and the effectiveness of the use of contextual descriptors was analyzed based on works [26, 27].

The greatest increase is observed in Privacy (+7.04%), Responsibility (+5.59%), and Freedom & Autonomy (+5.35%) indicators. This indicates increased attention to data protection, clear definition of AI responsibility, and ensuring user autonomy. Transparency (+2.56%), Justice & Fairness (+3.24%), and Non-maleficence (+3.70%) indicators also showed significant growth, indicating stricter requirements for AI system transparency and the avoidance of biases.

The overall trend of increasing all indicators, including those that initially had lower values (Sustainability +4.01%, Dignity +3.50%, Solidarity +3.60%), indicates an expanded understanding of AI's role in addressing social and environmental issues. This emphasizes the importance of considering a wide range of ethical aspects when developing and implementing AI systems in various fields and social contexts.

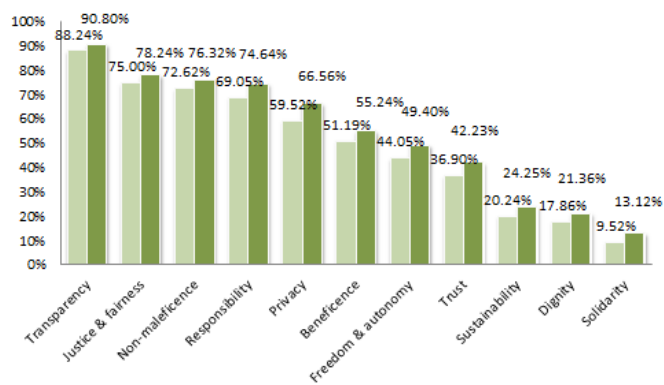

Fig. 3. Change in indicators due to the expansion of descriptor specification through the addition of contextual descriptors.

Indicators in this group (Transparency, Responsibility, Trust) demonstrate a high level of importance, particularly Transparency (90.80%). This highlights the critical role of AI system understandability and accountability.

In the Justice and Ethics group, Justice & Fairness (78.24%) and Non-maleficence (76.32%) have high indicators, indicating the priority of ethical aspects in AI development and application. The significant increase in Privacy (66.56%) reflects the increased attention to data protection. Freedom & Autonomy (49.40%) and Dignity (21.36%) indicators also increased, highlighting the importance of user rights.

Although Sustainability (24.25%) and Solidarity (13.12%) indicators remain relatively low, their increase indicates heightened attention to AI's role in addressing global challenges.

## VI. DISCUSSION

The analysis of changes in indicators demonstrates the significant impact of adding contextual descriptors to essential descriptors in knowledge representation and ontology alignment studies.

The most substantial increase is observed in the indicators of Privacy, Responsibility, and Freedom & Autonomy. This reflects the growing importance of data protection, clear definition of AI responsibility, and ensuring user autonomy in various contexts of AI technology application. In particular, the Privacy indicator increased by 7.04%, which is the largest increase among all indicators. This may be due to increased attention to data protection in sensitive areas such as healthcare and finance. Transparency, Justice & Fairness, and Non-maleficence indicators also showed significant growth. This indicates stricter requirements for AI system transparency and accountability, as well as increasing attention to ensuring equality and avoiding biases.

The limitations of the proposed method lie in its application in areas where context is insignificant or unchanging, such as highly formalized and strict knowledge systems, knowledge bases with well-defined ontologies, where only formal connections and definitions are relevant and sufficient for integrating contextual information. However, in most real-world cases, the integration of contextual descriptors can significantly enhance the accuracy and relevance of ontology alignment.

It is also interesting to note that indicators that initially had lower values, such as Sustainability, Dignity, and Solidarity, also showed noticeable growth. This may indicate an expanded understanding of AI's role in addressing social and environmental issues, as well as increased attention to respecting human dignity and social cohesion in the context of AI technology development.

It is also important to note that despite the overall increase, the relative order of indicators remained almost unchanged. This may indicate the stability of the main ethical priorities in the AI field, but with a deeper understanding of their manifestation in different contexts.

## VII. CONCLUSIONS

In this study, a new approach to ontology alignment was developed and experimentally tested, integrating contextual descriptors to improve the accuracy of semantic correspondence. The proposed approach allows for a more accurate consideration of the multidimensionality of knowledge, ensuring its relevant use in different contexts. Experimental results showed an improvement in metrics compared to the baseline method. Specifically, the average overall improvement in indicators was approximately 4.36%.

The method may be limited in areas with insignificant or unchanging context, where formal links in ontologies are sufficient. However, in most cases, contextual descriptors can significantly enhance the accuracy of ontology alignment.

These results confirm that the integration of contextual descriptors significantly improves the system's ability to consider nuances of semantic heterogeneity, especially when working with heterogeneous data and integrating information from various sources. The proposed approach opens new perspectives for creating flexible and adaptive knowledge systems that can effectively operate in complex and diverse conditions.

Future research can focus on optimizing weight coefficients and developing methods for automatic parameter tuning for different domains, allowing even more effective use of contextual descriptors in ontology alignment.